\documentclass{article}

\PassOptionsToPackage{numbers,compress}{natbib}

\usepackage[preprint]{neurips_2026}

\usepackage[utf8]{inputenc} 
\usepackage[T1]{fontenc}    
\usepackage{hyperref}       
\usepackage{url}            
\usepackage{booktabs}       
\usepackage{amsfonts}       
\usepackage{nicefrac}       
\usepackage{microtype}      
\usepackage[table]{xcolor}         
\usepackage{graphicx}
\usepackage{amssymb}
\usepackage{tikz}
\usetikzlibrary{calc}

\usepackage{tabularx}
\usepackage{amsmath}
\usepackage{array}
\usepackage{multirow}
\usepackage{enumitem}
\usepackage{ragged2e}
\usepackage{caption}
\usepackage{adjustbox}
\usepackage[skins,breakable]{tcolorbox}

\usepackage{tikz}

\newcolumntype{L}[1]{>{\raggedright\arraybackslash}p{#1}}
\newcolumntype{Y}{>{\raggedright\arraybackslash}X}
\newcommand{\tabhead}[1]{\textbf{#1}}
\newtcolorbox{takeawaybox}[1][Takeaway]{%
enhanced,
breakable,
colback=red!3, 
colframe=red!60!black, 
colbacktitle=red!70!black,
coltitle=white,
fonttitle=\bfseries,
boxrule=0.6pt,
boxsep=4pt,
left=0pt, right=0pt, top=0pt, bottom=0pt,
sharp corners,
halign=flush left,
}
\newcommand{\generatedfigure}[3]{%
  \IfFileExists{#1}{%
    \includegraphics[width=\linewidth,height=#2,keepaspectratio]{#1}%
  }{%
    \fbox{%
      \begin{minipage}[c][#2][c]{0.94\linewidth}
        \centering
        \small #3\\[2pt]
        \footnotesize Missing figure asset: \texttt{\detokenize{#1}}
      \end{minipage}%
    }%
  }%
}

\title{Beyond Perplexity: A Behavioral Evaluation Framework for Deployment-Memory Claims in LLM Test-Time Training}

\author{%
    Xiangchen Song$^{1}$ \quad
    Zhenhao Chen$^{2}$ \quad
    Lingjing Kong$^{1}$ \quad
    Shaoan Xie$^{1}$ \\ 
    \textbf{Xinshuai Dong}$^{1}$ \quad
    \textbf{Guangyi Chen}$^{1,2}$ \quad
    \textbf{Kun Zhang}$^{1,2}$ \\
    $^1$Carnegie Mellon University \quad $^2$MBZUAI\\
    \texttt{xiangchs@cs.cmu.edu, kunz1@cmu.edu} \\
}

\begin{document}

\maketitle

\begin{abstract}
Large language model test-time training (TTT) is often evaluated through local proxy metrics: models are updated on recent tokens, retrieved context, target-domain data, or verifiable task attempts, and then judged by perplexity, future-token loss, long-context performance, or reward. These metrics are well matched to claims about stream adaptation, domain adaptation, context compression, and reward-backed test-time improvement. They are weaker evidence, however, for a capability that TTT results are increasingly used to motivate: deployed assistant memory, personalization, or sparse post-deployment learning, which instead requires behavioral evidence such as later recall, paraphrase robustness, retention, locality, conflict handling, and use in downstream actions after the original support context is removed. We introduce a behavioral evaluation framework that calibrates TTT memory claims to the evidence that supports them. It has two components: a claim-calibrated evidence ladder that separates stream/domain adaptation, bridge internalization, and deployment-time behavioral learning; and an evaluation protocol with matched explicit-memory baselines and mutually exclusive failure categories. We validate the framework by auditing recent TTT and memory-adjacent work and by instantiating it as a controlled diagnostic in which, in a sparse nonce-fact setting, one-step LoRA updates lower support and answer loss across three Qwen3 model scales while generated free-form recall stays at zero, exposing a measurable gap between proxy improvement and deployment behavior. The framework gives authors and evaluators a concrete standard for aligning TTT memory claims with the evidence actually reported.

\end{abstract}

\section{Introduction}

Test-time training (TTT) challenges the conventional ``train, then deploy'' boundary by allowing model states or parameters to change during inference. In large language models, recent work has made this idea technically concrete: models may update from retrieved neighbors \citep{hardt2024tttnn}, learn online hidden-state updates through fast weights \citep{sun2024tttlayers}, perform large-chunk updates for throughput and state capacity \citep{zhang2025lact}, or align online updates with next-token prediction \citep{feng2026inplace}. Related work studies targeted context-specific updates \citep{bansal2025notjustcontext}, meta-learned long-context learning \citep{tandon2025e2e}, parameter-efficient context memories \citep{chen2026perk}, locally supported parametric memories \citep{lu2026locas}, input-perplexity minimization \citep{hu2025tlm}, label-free uncertainty signals \citep{xu2026sytta}, unlabeled reinforcement learning \citep{zuo2025ttrl}, and self-directed update data \citep{zweiger2025seal, acikgoz2025ttsi}.

Across these settings, the evaluation recipe is relatively stable. A model is updated at test time on recently observed tokens, retrieved examples, task attempts, or generated data, and performance is then reported through lower perplexity, better future-token prediction, improved long-context accuracy, or higher reward. These results show real progress on in-sequence adaptation and compact use of recent context, often addressing practical limits of transformer-based systems such as finite context windows or static parameters. They also explain why TTT is attractive for LLM systems: online updates may allow a model to adapt to new evidence rather than relying only on fixed parameters or the current prompt.

The central observation of this paper is that this evidence is not always calibrated to the claims it is used to support. In real-world assistant settings, the relevant question is often not whether an update improves prediction on a nearby continuation. It is whether a deployed model can hear a sparse user preference, acquire a project-specific fact, revise a stale belief, or learn a procedure, and then use that information later after the original support context is removed.

\textbf{We formalize this as a behavioral evaluation framework for LLM test-time training: claims about memory, personalization, or sparse post-deployment learning are calibrated against behavioral evidence beyond perplexity---later recall, paraphrase robustness, retention, locality, and conflict handling after the original support context is removed---rather than against local likelihood alone.}

The fact that perplexity is an imperfect proxy for downstream behavior is not new; similar lessons appear in model editing, retrieval, and memory evaluation. The distinctive issue for LLM TTT is that inference-time parameter or state updates make local likelihood gains especially easy to reinterpret as evidence of learning, memory, or personalization. We call this failure mode \emph{evidence migration}: evidence that directly supports one evaluation regime is carried into a stronger deployment narrative without the behavioral tests required for that narrative.

\begin{figure}[h]
\centering
\generatedfigure{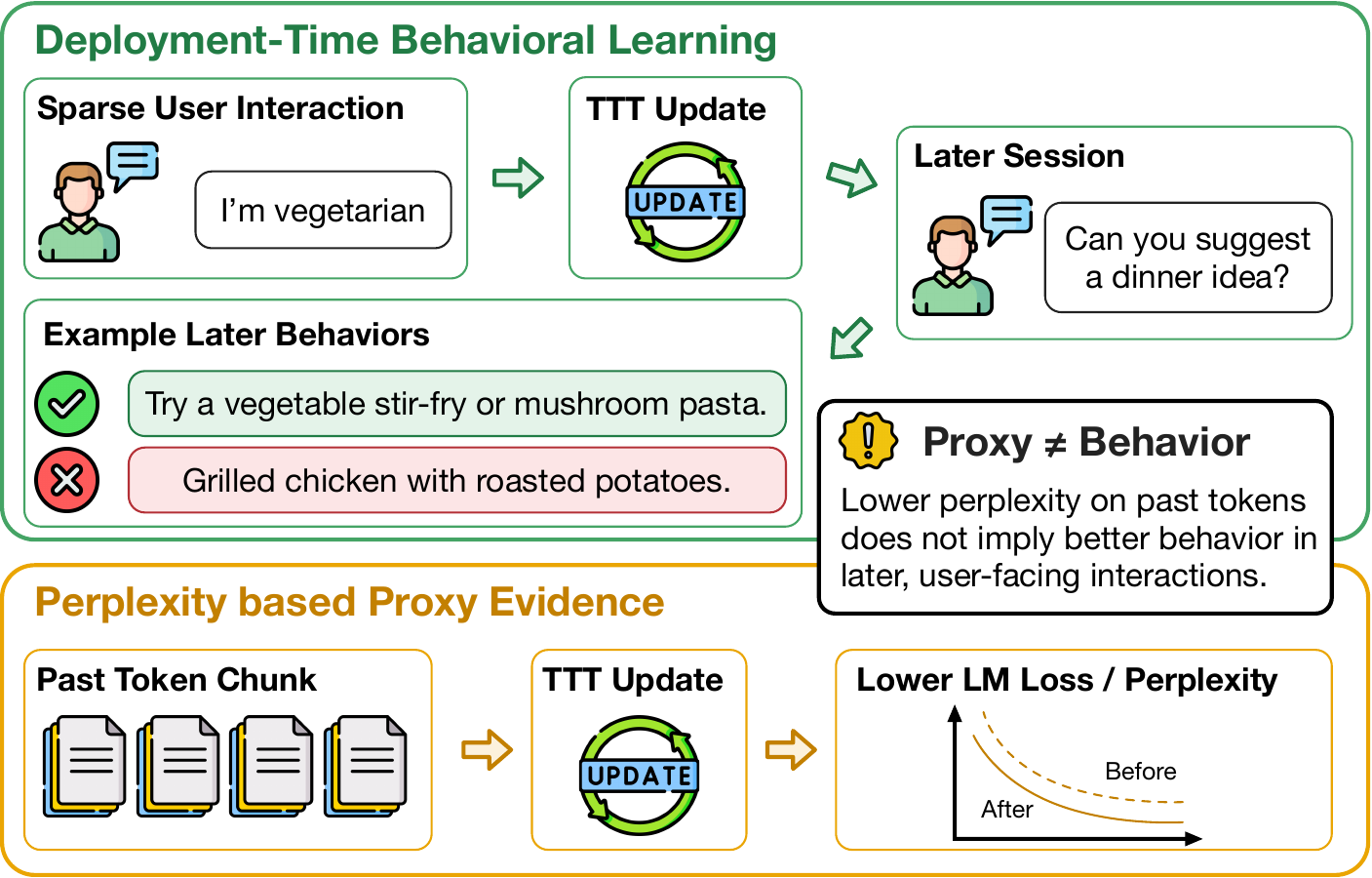}{2.7in}{Figure 1 image slot: deployed-assistant use case and TTT evaluation gap}

\caption{Two evaluation paradigms. Top: deployment-time behavioral learning from a sparse user interaction, illustrated with a user-stated dietary preference and evaluated by whether that preference changes later user-facing behavior. Bottom: perplexity-based evidence, where the model updates on a past support token chunk and achieves lower loss on a future chunk.}
\label{fig:regime-split}
\end{figure}

The key distinction is between \emph{in-sequence adaptation} and \emph{deployment-time behavioral learning}. In in-sequence adaptation, a model observes a token stream, updates its state or parameters on a support chunk, and is evaluated on future tokens from the same stream. In deployment-time behavioral learning, a deployed model receives sparse but high-value interactions---facts, preferences, corrections, or procedures---and must later use them in a personalized and stable way. Local loss reductions can strongly support the first setting while leaving the second only partially tested. Figure~\ref{fig:regime-split} illustrates this split with a sparse user-stated preference that should influence a later response.

This distinction matters because downstream systems are beginning to draw on recent TTT work for memory-like capabilities. Many long-context, label-free adaptation, and reward-backed TTT methods primarily report perplexity, likelihood, uncertainty, throughput, long-context accuracy, or task reward \citep{sun2024tttlayers, zhang2025lact, feng2026inplace, bansal2025notjustcontext, tandon2025e2e, hu2025tlm, xu2026sytta, zuo2025ttrl}. A closer frontier studies parameter-efficient context memories, self-adaptation, and context internalization \citep{chen2026perk, wang2024memoryllm, lu2026locas, zhang2026absorber}. These are important steps toward memory-like systems, but deployment-time memory adds a further requirement: sparse user information should remain usable after delay and should be evaluated against retrieval or long-context baselines.

Many existing TTT papers are properly grounded in proxy evaluation and do not themselves claim to solve deployment-time memory. The problem arises when such results are later cited, summarized, or motivated as evidence for stronger claims about memory, personalization, or self-updating assistants. Table~\ref{tab:evidence-migration-patterns} summarizes common evidence migration patterns and the behavioral tests needed to support the stronger deployment-time narratives.

\begin{table}[t]
\caption{Evidence migration patterns. Evidence that is well grounded for one claim can be used to motivate stronger deployment-time narratives that require additional behavioral tests.}
\label{tab:evidence-migration-patterns}
\centering
{\scriptsize
\setlength{\tabcolsep}{2.6pt}
\renewcommand{\arraystretch}{1.08}
\begin{tabularx}{\linewidth}{@{}L{0.20\linewidth} Y Y Y@{}}
\toprule
\tabhead{Pattern} & \tabhead{Grounded evidence} & \tabhead{Deployment-time claim} & \tabhead{Missing behavioral evidence} \\
\midrule
\textbf{Token-stream TTT} & lower future-token loss or long-context task gain & the model learns user facts online & sparse fact recall after context removal, paraphrase, and delay \\
\midrule
\textbf{Context compression / internalization} & needle recall, context ablation, or parametric-memory evidence & forms persistent assistant memory & cross-session recall, locality, and conflict behavior \\
\midrule
\textbf{Self-adaptation from generated data} & task improvement under generated updates & personalized continual learning & weak user evidence, user-specific retention, and correction \\
\midrule
\textbf{Reward-rich test-time discovery} & verifiable reward improvement & broad deployment self-improvement & reward-poor user facts, preferences, and procedures \\
\bottomrule
\end{tabularx}
}
\end{table}

We use TTT as an umbrella term for inference-time methods that update model state, especially parameters or fast weights, and treat LLM \emph{test-time learning} (TTL) as a neighboring adaptation formulation. We use \emph{memory}, \emph{continual learning}, and \emph{deployment-time behavioral learning} for the stronger claim that new information encountered after deployment can be acquired and later used. When we discuss \emph{future-token gain}, we mean improvement on nearby held-out continuations after a test-time update.

Concretely, this paper makes four contributions:

\begin{enumerate}[leftmargin=20pt]

    \item We identify and name \emph{evidence migration}: the failure mode in which TTT results grounded in perplexity, reward, or same-stream adaptation are used to support stronger claims about memory, personalization, or post-deployment learning.

    \item We introduce a claim-calibrated evidence ladder, ranging from proxy evidence for stream adaptation, to evidence for context internalization, to behavioral evidence for deployment-time memory, and pair it with an evaluation protocol and matched explicit-memory baselines.

    \item We audit recent TTT and memory-adjacent work through this ladder, distinguishing what current evaluations directly support from what additional tests would be needed for stronger deployment-time claims.

    \item We instantiate and stress-test the framework with a controlled Qwen3/LoRA diagnostic, showing that proxy and answer-likelihood gains can coexist with zero generated recall, and provide claim-specific behavioral templates for factual, preference, correction, procedure, and agent-memory claims.

\end{enumerate}


\section{Perplexity is an Incomplete Proxy}

Perplexity remains valuable. It is often the cleanest early signal that a test-time update has changed the model in a nontrivial way. The problem is not that perplexity is uninformative, but that it answers a narrower question than many deployment-time claims require. Lower loss can arise from mechanisms that are useful for stream adaptation but insufficient for post-deployment learning.

\textbf{Continuous text is a dense-evidence regime.}
When support and evaluation chunks come from the same passage, they share topic, entities, lexical choices, and local discourse structure. Lower future-token loss may therefore reflect continuation fitting, topic priming, or memorization of nearby regularities, which is precisely what dynamic evaluation \citep{krause2018dynamic, rannentriki2024dynamic} was designed to exploit. This is a legitimate stream-adaptation benefit. However, it does not require the model to form a stable, queryable representation of the underlying fact. Dense same-stream evidence is therefore weaker evidence for sparse deployment-time learning.

\textbf{Associative retrieval differs from semantic acquisition.}
Long-context TTT can also succeed by building efficient associative state for the current input. Some long-context TTT results are consistent with the view that TTT-style state acts as a compressed associative memory over the current context. For example, TTT-E2E \citep{tandon2025e2e} connects TTT-style updates to key-value binding mechanisms and reports that full attention remains substantially stronger on needle-in-a-haystack recall, partly because compressed TTT-style state can discard details needed for exact retrieval. This supports context compression. It is weaker evidence for acquiring a reusable fact, rule, preference, or procedure that should survive paraphrase and delay.

\textbf{Teacher forcing is easier than behavioral extraction.}
A model can assign higher probability to the gold continuation of a support fact without later producing the correct answer to a free-form query about that fact. This gap is especially important under open-ended decoding: a small increase in the probability of the right answer tokens may still be insufficient for greedy generation, beam search, or robust answer selection under paraphrase. Lower teacher-forced loss is therefore evidence that the update moved probability mass in a useful direction, but it does not guarantee that the learned information is accessible during open-ended generation. A system can improve language-model loss while still failing to retrieve the new knowledge behaviorally.

\textbf{Cumulative stream adaptation can be mistaken for sparse-interaction learning.}
In standard continuous-text TTT protocols, updates often accumulate across a long input stream. This is appropriate when the goal is online adaptation to that stream. It becomes harder to interpret when the same result is used as evidence that a single sparse user interaction can be internalized and reused later. A stream protocol can conflate immediate adaptation to the current support item with cumulative adaptation from all preceding chunks. Deployment-time claims therefore need reset-vs-stream comparisons, or equivalent controls, when the protocol permits them.

\textbf{Local LM metrics do not test generalization, locality, or conflict.}
Deployment-time learning requires more than improved prediction of nearby text. The update should persist under paraphrase, delay, mild distribution shift, and conflicting later evidence, while leaving unrelated behavior intact. The model editing literature provides a useful cautionary parallel: ROME \citep{meng2022rome} and MEMIT \citep{meng2023memit} made efficacy, generalization, and specificity explicit, while later evaluations found that short-form edit tests can miss specificity failures \citep{hoelscherobermaier2023detecting}, multi-hop failures \citep{zhong2023mquake}, long-form failures \citep{rosati2024leme}, gradual and catastrophic forgetting under scale \citep{gupta2024editing-scale}, broader evaluation mismatches for edited models \citep{li2024shouldedit}, and deployment-style failures in the wild \citep{yang2025mirage}. The same lesson carries over to TTT memory claims: a locally valid success metric can overstate the practical capability of interest when the claim concerns delayed, user-facing behavior.

\begin{takeawaybox}
\textbf{Takeaway:} Support reconstruction, future-token loss, answer NLL, and, when used, candidate ranking should remain in TTT reports because they are informative mechanism-level signals. For deployment-time TTT, however, they should be labeled as \emph{proxy metrics} and kept separate from headline behavioral evidence.
\end{takeawaybox}

\section{Deployment-Time Memory as a Behavioral Evaluation Target}

This limitation does not invalidate existing TTT evaluations for stream adaptation, long-context compression, domain adaptation, or reward-backed test-time improvement. The issue is claim calibration: these evaluations do not by themselves establish the stronger deployment-time capabilities implied by memory, personalization, or self-updating assistant narratives.

The regime split matters because memory-style deployment is no longer an abstract desideratum. It is already the target of direct behavioral evaluation. LoCoMo \citep{maharana2024locomo} evaluates very long-term conversational memory; LongMemEval \citep{wu2024longmemeval} benchmarks long-term interactive memory in chat assistants; MemoryAgentBench \citep{hu2025memoryagentbench} evaluates memory through incremental multi-turn interactions; MemoryBench \citep{ai2025memorybench} targets memory and continual learning in LLM systems; MemoryCD \citep{zhang2026memorycd} studies lifelong cross-domain personalization; and Mem2ActBench \citep{shen2026mem2actbench} evaluates long-term memory utilization in task-oriented agents. These benchmarks ask whether information from one or a few interactions can be reused after delay, paraphrase, abstention pressure, or conflict. Improving nearby continuation loss after a dense support chunk answers a different question.

Deployment-time learning differs from pretraining and continuous-text adaptation in three ways. It is \emph{low redundancy}: a fact, preference, correction, or procedure may be stated only once. It is \emph{weak and heterogeneous}: user evidence often arrives as short, informal utterances rather than clean supervised examples. It is also \emph{delayed and behavioral}: the model must later answer, revise, abstain, route, or act according to the learned information without damaging unrelated behavior. These properties make behavioral probes necessary: the central question is whether the update changes later behavior in the way required by the claimed deployment use case, not merely whether it improves local likelihood.

The distinction is especially clear in settings with stronger supervision. TTT-Discover \citep{yuksekgonul2026tttdiscover} shows that test-time training can improve search when the environment supplies verifiable rewards and many attempts can be scored. Personalized assistants face a harder frontier: semantic internalization under weak evidence, as explored by SEAL \citep{zweiger2025seal} and Absorber LLM \citep{zhang2026absorber}, and stressed by direct memory benchmarks such as MemoryBench \citep{ai2025memorybench}. Reward-rich discovery and dense stream adaptation are promising, but sparse deployment-time memory imposes a separate behavioral burden.

Matched explicit-memory baselines are therefore central. If the deployed task is to remember a user fact, preference, correction, or procedure, then retrieval, long-context prompting, and memory systems that store and reuse the same evidence are natural comparators. MemoryBank \citep{zhong2024memorybank} stores long-term user memory, LongMem \citep{wang2023longmem} augments language models with long-term memory, MemGPT \citep{packer2023memgpt} manages memory through an operating-system-like architecture, Mem0 \citep{chhikara2025mem0} targets scalable production-ready long-term memory for agents, Dynamic Cheatsheet \citep{suzgun2026dynamiccheatsheet} maintains an adaptive test-time memory, and MEMORYLLM \citep{wang2024memoryllm} studies self-updatable language models. These systems retain deployment-time information outside ordinary base-model weights and retrieve, page, curate, or pool it when needed. They need not be framed as competitors that TTT must always beat. Rather, they define the behavioral and systems-level tradeoff that a parametric update must justify.

\begin{takeawaybox}
\textbf{Takeaway:} Deployment-time memory claims require behavioral tests against matched explicit-memory baselines. These comparisons reveal whether parametric TTT adds value under privacy, latency, compression, or context-pressure constraints.
\end{takeawaybox}

\section{Audit}

\begin{figure}[h]
\centering
\includegraphics[width=0.92\linewidth]{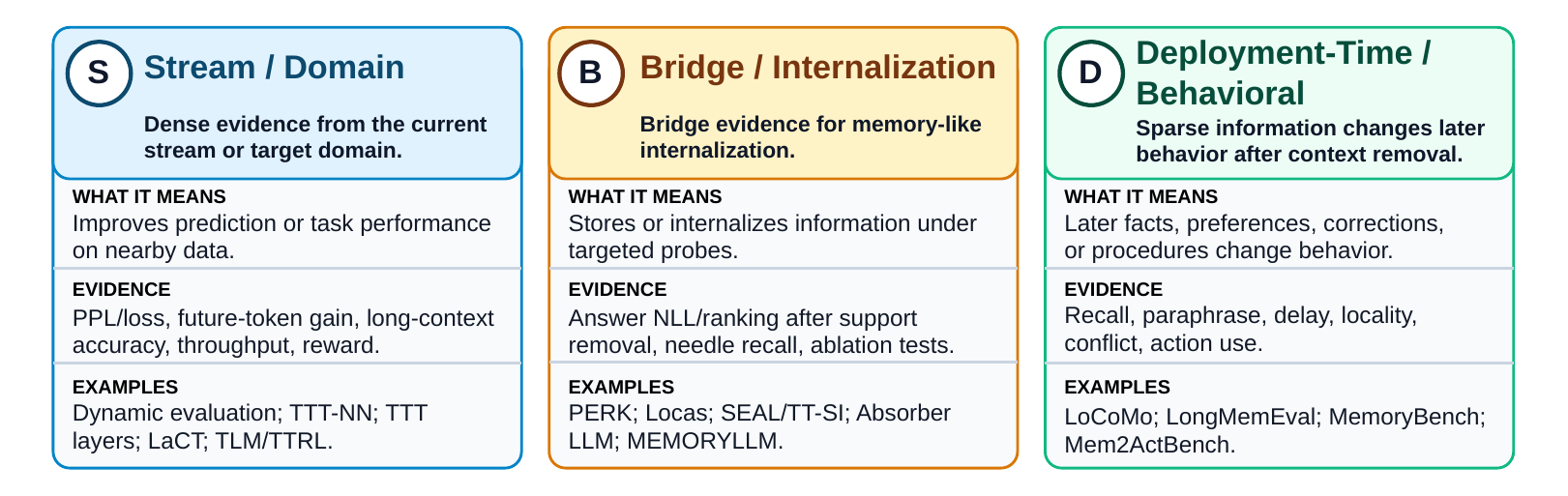}
\captionsetup{type=figure,hypcap=false}
\caption{Three claim levels used to calibrate TTT claims. The level records the strongest evidence directly supported by the reported evaluation, not the broadest motivation or downstream narrative.}
\label{fig:claim-levels}
\vspace{-0.5em}
\end{figure}

\textbf{Claim levels.}
We audit recent TTT and memory-adjacent work using three claim levels. \emph{S-level} evidence denotes stream or domain adaptation: the model is updated on a recent stream, target-domain data, retrieved examples, or task attempts, and evaluated by nearby loss, same-stream performance, domain accuracy, long-context performance, or reward. \emph{B-level} evidence denotes bridge mechanisms such as internalization, parametric memory, context absorption, or self-adaptation: the evaluation suggests that information can be stored, compressed, or transformed inside model state, but does not yet establish sparse delayed user-facing behavior. \emph{D-level} evidence denotes deployment-time behavioral learning: sparse post-deployment information changes later behavior under recall, paraphrase, delay, locality, conflict, or action-use tests after the original support context is unavailable. Figure~\ref{fig:claim-levels} summarizes the ladder.

Operationally, we assign S when the primary evidence is same-stream, domain, or task-adaptation performance; B when the evaluation tests memory-like internalization, parametric storage, context absorption, or self-adaptation without fully testing sparse delayed user behavior; and D only when the evaluation directly probes later behavior from sparse post-deployment information after context removal. B-level is intentionally heterogeneous: it covers mechanisms that bridge proxy adaptation and deployment memory, but these mechanisms do not by themselves establish D-level memory.

\textbf{Audit protocol.}
We screened more than 40 targeted-search candidates available through April 2026 using bibliography seeding and targeted queries around LLM TTT/TTL, long-context TTT, context memory, parametric memory, personalization, continual learning, and assistant memory. A record was included if its title, abstract, introduction, or motivation explicitly invoked test-time learning or training, memory, persistence, context absorption, self-improvement, personalization, continual learning, or assistant memory. We coded 24 papers and treated the remaining records as background or exclusions, including retrieval-only or external-memory baselines, model-editing and safety background, non-LLM TTT, and architecture-adjacent context memory.

We therefore code claims by calibrating them to evidence rather than by inferring intent, assigning each paper the strongest claim level directly supported by its reported evaluations instead of the level suggested by its broadest motivating statement. Assistant-memory benchmarks are labeled as D targets rather than D-level TTT evidence because they define the behavioral task family but do not themselves show that parametric TTT achieves it. This audit is not a prevalence estimate or systematic review. Appendix~\ref{app:audit-sheet} gives the search fields, candidate accounting, background and exclusion summary, and boundary-case sensitivity worksheet.

The audit asks what each paper family's evidence directly supports, and what additional tests would be needed before using it in a deployment-memory claim. Table~\ref{tab:mismatch-exemplars} gives representative examples.

\begin{table}[h]
\vspace{-1em}
\caption{Claim-calibration exemplars. The level column records the strongest claim level directly supported by the reported evaluation, not the broadest motivation or downstream narrative. Full paper-level coding and boundary cases are in Appendix~\ref{app:audit-sheet}.}

\label{tab:mismatch-exemplars}
\centering
{\small
\setlength{\tabcolsep}{2.5pt}
\renewcommand{\arraystretch}{1.05}
\begin{tabularx}{\linewidth}{@{}L{0.22\linewidth} L{0.12\linewidth} Y Y@{}}
\toprule
\tabhead{Paper family} & \tabhead{Level} & \tabhead{Evidence directly supported} & \tabhead{Needed before D-level deployment-memory use} \\
\midrule
TTT layers / LaCT / In-Place TTT \citep{sun2024tttlayers, zhang2025lact, feng2026inplace} & S & Test-time state improves stream, domain, or long-context modeling under dense evidence & Sparse fact or preference behavior after context removal, paraphrase, delay, and locality probes \\
\midrule
PERK / Locas / Absorber LLM \citep{chen2026perk, lu2026locas, zhang2026absorber} & B & Parameter-efficient memories, local parametric support, or context absorption provide bridge evidence & One-shot facts, preferences, corrections, conflict overwrite, locality, and matched retrieval baselines \\
\midrule
SEAL / TT-SI / TTT-Discover \citep{zweiger2025seal, acikgoz2025ttsi, yuksekgonul2026tttdiscover} & B & Self-adaptation, agent improvement, or reward-backed discovery under generated or verifiable update data & Weak user evidence, user-specific retention, correction, and reward-poor deployment settings \\
\midrule
Assistant-memory benchmarks \citep{wu2024longmemeval, ai2025memorybench, shen2026mem2actbench} & D target & Direct behavioral tasks for long-term assistant memory, corrections, personalization, or actions & Matched TTT update budget, explicit-memory baseline, and no-context ablations before using benchmark success to justify parametric TTT claims \\
\bottomrule
\end{tabularx}
}
\vspace{-0.5em}
\end{table}

\textbf{Audit pattern.}
The main pattern is not that TTT lacks progress; it is that most reported evidence remains below D-level deployment behavior. Dynamic evaluation, TTT-NN, TTT layers, LaCT, In-Place TTT, TTT-E2E, TLM, SyTTA, and TTRL primarily provide S-level evidence: adaptation to recent text, target-domain data, or verifiable tasks. PERK, Locas, SEAL, TT-SI, TTT-Discover, MEMORYLLM, and Absorber LLM move toward B-level evidence by studying parametric memories, self-adaptation, context internalization, or reward-backed discovery. Reasonable reclassifications of S/B boundary cases may change family counts, but they do not change the decision rule. A D-level deployment claim requires behavioral tests showing that a sparse user fact, preference, correction, or procedure changes later behavior after paraphrase, delay, conflict, and removal of the original context.

\begin{takeawaybox}
\textbf{Takeaway:} S- and B-level evidence can motivate deployment-memory hypotheses, but D-level language requires behavioral tests for sparse information after context removal, paraphrase, delay, locality, and conflict.
\end{takeawaybox}

\section{A Controlled Diagnostic Experiment}
\label{sec:controlled-diagnostic}

We instantiate the framework with a controlled diagnostic experiment that isolates a common inference behind evidence migration: whether improved support loss or answer likelihood is sufficient evidence for deployment recall. A test-time update is applied to a sparse factual support sentence, the support context is then removed, and the model is later queried for the injected fact. Such a cleanly controlled case, in which proxy metrics improve while later behavior does not, demonstrates why proxy evidence and deployment recall should be reported separately.

\textbf{Diagnostic setup.}
We use Qwen3 models at three scales, \texttt{Qwen/Qwen3-1.7B}, \texttt{Qwen/Qwen3-4B}, and \texttt{Qwen/Qwen3-8B}, with LoRA adaptation and a minimal online update on each support text. The factual setting introduces nonce access-code facts and then tests direct, paraphrased, and delayed recall after the support context is removed. We define $\Delta\mathrm{NLL}=\mathrm{NLL}_{\mathrm{after}}-\mathrm{NLL}_{\mathrm{before}}$, so negative values indicate improved teacher-forced loss after the update. Full hyperparameters, prompt counts, scoring rules, explicit-memory baselines, conflict-overwrite tests, stress updates, and robustness checks are in Appendices~\ref{app:evaluation-agenda} and~\ref{app:diagnostic-robustness}.

\textbf{Evidence levels.}
The diagnostic keeps proxy, bridge, and target behavioral evidence separate. \emph{Proxy} evidence consists of support reconstruction and local $\Delta$NLL. \emph{Bridge} evidence consists of answer $\Delta$NLL. This metric tests whether probability mass moves toward the target answer, but not whether the model will produce that answer in open-ended use. \emph{Target behavior} in the main factual probe is generated success under direct, paraphrased, and delayed prompting after the support context is removed. Appendix~\ref{app:diagnostic-robustness} adds locality, conflict-overwrite, retrieval, and replacement-memory checks as robustness and comparison evidence. This separation suggests lower loss is informative, but delayed free-form recall must be measured behaviorally.

\begin{table}[h]
\vspace{-1em}
\caption{Loss improves, but generated free-form recall does not appear under the fixed greedy decoding protocol. Values are percentages except $\Delta$NLL. Lower $\Delta$NLL indicates improved teacher-forced loss after the update. Generated success is measured over 48 factual probes per model using normalized first-answer-line containment of the target answer.}
\label{tab:update-loss-recall}
\centering
{\small
\setlength{\tabcolsep}{4.2pt}
\renewcommand{\arraystretch}{1.08}
\begin{tabular}{@{}lccc ccc@{}}
\toprule
& \multicolumn{3}{c}{Loss improvement $\Delta$NLL $\downarrow$}
& \multicolumn{3}{c}{Greedy generated success (\%) $\uparrow$} \\
\cmidrule(lr){2-4}\cmidrule(lr){5-7}
Model & Support & Direct answer & Paraphrased answer & Direct & Paraphrased & Delayed \\
\midrule
Qwen3-1.7B & $-1.151{\pm}0.051$ & $-0.674{\pm}0.085$ & $-0.714{\pm}0.071$ & $0.0$ & $0.0$ & $0.0$ \\
Qwen3-4B   & $-1.218{\pm}0.093$ & $-0.529{\pm}0.081$ & $-0.531{\pm}0.071$ & $0.0$ & $0.0$ & $0.0$ \\
Qwen3-8B   & $-0.966{\pm}0.026$ & $-0.504{\pm}0.079$ & $-0.463{\pm}0.070$ & $0.0$ & $0.0$ & $0.0$ \\
\bottomrule
\end{tabular}
}
\vspace{-0.5em}
\end{table}

\textbf{Results.}
Across all three Qwen3 sizes, one-step LoRA updates improve support and answer loss, while generated free-form recall remains at 0.0\% for direct, paraphrased, and delayed queries under the fixed greedy decoding and normalized-containment scoring protocol. Table~\ref{tab:update-loss-recall} shows this proxy/behavior split directly, support $\Delta$NLL improves substantially, and answer $\Delta$NLL also improves, yet these gains do not translate into successful greedy recovery of the injected facts.

The zero generated-recall result is not the only signal we report. The same table shows that teacher-forced answer likelihood improves, yet greedy generation still omits the target fact. Appendix~\ref{app:diagnostic-robustness} gives representative cases where the likelihood signal does not translate into generated recall.

\textbf{Explicit-memory controls.}
Matched explicit-memory controls serve as evidence-usability and deployment-comparison checks. The appendix specifies exact-context, BM25-style retrieval, and replacement-memory protocols, but the factual aggregate control reported here is the easy BM25 condition. In the 1.7B seed-replication checks, BM25 hit and memory answering remain at $48/48$. Harder retrieval checks in Appendix~\ref{app:diagnostic-robustness} have Hit@1 $0/48$ under paraphrased-support decoys and $0/24$ under stale/current conflicts. Thus, the point is not that retrieval trivially solves deployment memory, but that parametric TTT should be compared to explicit memory under matched evidence, query, scoring, and budget constraints.

\textbf{Stress update and conflict checks.}
A stronger update can produce generated recall, but it exposes the missing locality dimension. In the Qwen3-1.7B factual setting, the appendix stress check raises direct and delayed recall to 72.9\% ($35/48$) and paraphrased recall to 54.2\% ($26/48$), while locality falls from 97.9\% ($141/144$) to 9.7\% ($14/144$) (Table~\ref{tab:stress-update-tradeoff}). Conflict-overwrite checks show a different failure mode: the one-step LoRA update returns neither stale nor corrected code in all $72/72$ conflicts, while replacement memory is corrected-only in $68/72$ cases and both-corrected-and-stale in $4/72$ (Table~\ref{tab:conflict-overwrite-categories}). These checks show why D-level evidence should jointly report recall, paraphrase robustness, delay, locality or interference, conflict behavior, and matched baselines.

\textbf{Robustness scope.}
Appendix~\ref{app:diagnostic-robustness} reports additional checks, the 1.7B seed replications keep direct/paraphrased/delayed recall at $0/48$, preference/correction and procedure mini-tasks also improve proxy loss while leaving greedy behavior at $0/24$, and the continuous-text check is reported only as proxy-regime context. We therefore use these checks to qualify the diagnostic, not to claim that the tested one-step LoRA update achieves deployment memory.

\begin{takeawaybox}
\textbf{Takeaway:} The diagnostic separates proxy improvement from deployment recall. Proxy and bridge gains can coexist with zero behavioral recall, and stronger recall can expose severe locality cost. D-level claims, therefore, require generated recall under paraphrase and delay, locality, or conflict reporting, and matched baselines.
\end{takeawaybox}

\section{A Claim-Calibrated Evaluation Protocol}

For TTT results that invoke post-deployment learning, memory, personalization, or self-updating assistants, the framework maps claim language to evidence level. Table~\ref{tab:decisive} gives the decision rule for assigning supported wording from the reported evidence. Stream-adaptation claims can headline stream loss, future-token loss, long-context accuracy, throughput, or reward. Deployment-memory claims should headline later behavior under a deployment-like update episode.

\begin{table}[h]
\vspace{-1em}
\caption{Claim-calibrated decision protocol. Evidence can be positive and useful while still supporting only S- or B-level wording. D-level language requires behavioral improvement under the claimed use case, disclosed scoring, locality or failure reporting, and matched baseline comparison.}
\label{tab:decisive}
\centering
\scriptsize
\setlength{\tabcolsep}{2.3pt}
\renewcommand{\arraystretch}{1.08}
\begin{tabularx}{\linewidth}{@{}L{0.18\linewidth} L{0.24\linewidth} Y L{0.25\linewidth}@{}}
\toprule
\tabhead{Evidence reported} & \tabhead{Supported wording} & \tabhead{Requires more evidence} & \tabhead{Next evidence upgrade} \\
\midrule
\tabhead{Future-token loss, stream/domain metrics, or reward only} & Improves stream adaptation, target-domain adaptation, or task performance under the stated objective & Remembers user facts, learns preferences, or performs post-deployment continual learning & Reset-vs-stream controls and direct behavioral probes for the claimed deployment target \\
\midrule
\tabhead{Answer NLL / ranking, needle recall, context ablation} & Provides bridge evidence for compression or internalization under the tested protocol & Robust deployment memory, personalization, or self-updating assistant behavior & Direct generation, paraphrase, delay, and no-support-context evaluation \\
\midrule
\tabhead{Direct generated recall only} & Immediate write-in under a narrow prompt format & Persistent or robust memory, broad personalization, or safe correction & Paraphrased and delayed recall, locality, conflict categories \\
\midrule
\tabhead{Direct + paraphrase + delay + locality + conflict + matched baseline} & Deployment-memory claim under the stated setting and budget & Broad general memory beyond the tested setting, user population, or update mechanism & Broader tasks, stronger baselines, more seeds/models, and stated deployment constraints \\
\bottomrule
\end{tabularx}
\vspace{-1em}
\end{table}

\textbf{Match probes to claims.}
D-level evaluation should be claim-specific rather than maximalist. Factual memory requires no-context recall under paraphrase and delay; preference personalization requires later choices or response changes under plausible alternatives; correction claims require mutually exclusive stale/current scoring; procedural claims require later action or task transfer; and agent-memory claims require learned information to affect routing, tool calls, or arguments. Appendix~\ref{app:evaluation-agenda} gives operational templates and evidence tiers. Such evaluations can be small: the support episode should introduce a sparse item absent from the evaluation prompt, and the later probe should test the failure modes relevant to the claim. When possible, reset-vs-stream controls should separate one-shot write-in from accumulated stream adaptation.

\textbf{Keep proxy metrics separate.}
Support reconstruction, future-token loss, answer NLL, and, when reported, candidate ranking should remain in the report because they explain what the update is doing. But they should be presented as mechanism-level or intermediate signals, not as substitutes for the behavioral target. In particular, answer likelihood and generated answer success should be reported separately: the diagnostic above shows that a model can improve the likelihood of the target answer while still producing the same wrong free-form response.

\textbf{Use matched baselines.}
A no-update model tests whether the update helps at all. Exact-context or long-context prompting tests whether the support evidence is sufficient when explicit. Retrieval or memory baselines test whether the same evidence can be stored outside the weights and reused under the same query and scoring rule. A \emph{matched} baseline should use the same support information, query set, scoring rule, and disclosed budgets, including update tokens, trainable parameters, optimization steps, latency, memory footprint, context length, and retrieval index size when applicable.

Parametric TTT need not beat every explicit-memory baseline on every axis to be useful. Its value may appear under privacy, compression, latency, amortization, offline-operation, or context-pressure constraints. But those constraints should be stated and evaluated directly; otherwise, a memory or personalization claim risks conflating mechanism-level adaptation with deployment-time usefulness.

\begin{takeawaybox}
\textbf{Takeaway:} Claim language should track evidence level. Proxy and bridge metrics explain mechanisms, but memory, personalization, and post-deployment learning claims require behavioral evidence under the claimed use case, with failure categories and matched baselines visible.
\end{takeawaybox}

\section{Discussion}

The framework is deliberately scoped, and several natural objections and boundary cases arise about how far it should be applied. We address them here as design considerations that clarify what the framework does and does not require.

\noindent\textbf{Scope: evaluation against stated objectives.}
Many TTT papers study stream adaptation, long-context scaling, or task reward under the objective they optimize, without claiming to implement deployment-time memory. Those papers should be evaluated on their stated objectives. The interpretive issue arises when such results are later used to motivate memory, personalization, or self-updating assistants. At that point, S- and B-level evidence needs additional behavioral probes before it supports D-level language.

\noindent\textbf{Perplexity as behavior in dense streams.}
One response is that perplexity is already behavioral for language models, because next-token prediction is the model's core behavior. This is reasonable in dense-stream settings, where better prediction may be the desired outcome. Personalized assistants require a different kind of behavior: recalling sparse facts, applying corrections, resolving conflicts, or abstaining when information is missing. In that regime, lower loss is useful evidence about the update mechanism, but deployed memory still has to be shown in later responses.

\noindent\textbf{Relation to external-memory substrates.}
Assistant memory may often be better implemented explicitly rather than parametrically. Systems such as MemoryBank and MemGPT store, retrieve, page, or curate information outside the base-model weights, and they are aimed directly at long-term interaction and multi-session use \citep{zhong2024memorybank, packer2023memgpt}. The framework is compatible with this view. If explicit memory is the practical baseline, then parametric TTT should be evaluated against it under matched evidence, query, scoring, and budget constraints.

\noindent\textbf{Context compression as a legitimate target.}
Methods need not target sparse user facts to be memory-relevant. PERK, Locas, MEMORYLLM, and related systems study adapters, parametric memories, or persistent memory pools that encode context for later use \citep{chen2026perk, lu2026locas, wang2024memoryllm}. We treat this as a legitimate B-level objective. The calibration point is narrower: compression or internalization results should be described as such unless they also show deployment behavior under paraphrase, delay, conflict, and locality.

\noindent\textbf{Procedural rather than episodic memory.}
Test-time learning may be most useful for reusable strategies, code snippets, search heuristics, or reward-backed reasoning rather than user-fact recall. Dynamic Cheatsheet and TTRL are examples of this direction: they use accumulated solutions or reward-style signals to improve future problem solving \citep{suzgun2026dynamiccheatsheet, zuo2025ttrl}. The same evidence standard applies, but the behavioral target changes. A procedural-memory claim should be tested through later task performance, transfer, and failure modes for the learned procedure, rather than through factual recall alone.

\noindent\textbf{Stronger update mechanisms.}
Stronger TTT systems may pass behavioral tests that simple update mechanisms fail. Better objectives, routing, replay, constrained updates, verification, or hybrid parametric--external memory could produce useful deployment-time memory. The framework accommodates this possibility, and specifies what evidence would be needed to establish it: behavioral success under the claimed use case, together with update cost, locality, conflict, and matched baselines.

\section{Conclusion}

TTT has made real progress on long-context modeling, stream adaptation, domain adaptation, and reward-backed test-time improvement. Our framework provides a concrete way to keep claims and evidence aligned: under the proposed protocol, deployment-memory wording is supported only when the claimed deployment behavior is directly tested. Lower perplexity, future-token loss, answer likelihood, candidate ranking when explicitly reported, and reward under a stated task are valuable evidence for the objectives they measure, but they should not be treated as sufficient evidence for deployment-time memory.

Showing that a model has acquired a user fact, preference, correction, or procedure for later use requires tests that resemble the claimed use case: direct, paraphrased, and delayed behavior; locality and conflict reporting; disclosed update budgets; mutually exclusive failure categories; and matched retrieval or long-context baselines. The aim is to keep claim and evidence at the same level. If future TTT systems can provide memory, the decisive evidence will appear where memory matters: in later behavior, under realistic constraints, with the interference cost visible.



\bibliographystyle{plain}
\bibliography{ref}

\clearpage
\appendix
\setcounter{figure}{0}
\setcounter{table}{0}
\renewcommand{\thefigure}{A\arabic{figure}}
\renewcommand{\thetable}{A\arabic{table}}
\captionsetup[table]{hypcap=false}

\section{Evidence migration patterns}
\label{app:evidence-migration}

This appendix makes the evidence-migration concern concrete without treating the cited papers as overclaiming. The migration risk is interpretive: a result that is well scoped as S- or B-level evidence can become misleading when reused as support for a D-level deployment-memory narrative.

\begin{center}
\begin{minipage}{\linewidth}
\captionof{table}{Concrete evidence-migration patterns.}
\label{tab:appendix-evidence-migration-patterns}
\vspace{3pt}

\centering
{\footnotesize
\setlength{\tabcolsep}{3pt}
\renewcommand{\arraystretch}{1.08}
\begin{tabularx}{\linewidth}{L{2.35cm} Y Y Y}
\toprule
\tabhead{Migration pattern} & \tabhead{Grounded evidence} & \tabhead{Tempting stronger narrative} & \tabhead{Missing behavioral evidence} \\
\midrule
Dense-stream TTT $\rightarrow$ online user memory & Lower future-token loss or better long-context performance after updates on dense nearby text. & The model can learn new user facts online. & Sparse fact write-in, no-support-context recall, paraphrase, delay, and locality. \\
\addlinespace[1pt]
Context compression/internalization $\rightarrow$ persistent assistant memory & Needle recall, context ablation, parameter-efficient context state, or local parametric support. & The compressed state functions as durable assistant memory. & Cross-session or delayed behavior, conflict overwrite, unrelated-behavior preservation, and matched explicit memory. \\
\addlinespace[1pt]
Self-adaptation/generated updates $\rightarrow$ personalized continual learning & Task improvement after generated update data or self-improvement episodes. & The assistant can personalize from weak user evidence. & User-specific preference or correction probes, retention, overgeneralization checks, and safety-relevant conflict scoring. \\
\addlinespace[1pt]
Reward-backed discovery $\rightarrow$ broad deployment self-improvement & Test-time improvement in settings with verifiable rewards or many scored attempts. & The system can self-improve broadly after deployment. & Reward-poor user facts, preferences, procedures, held-out transfer, and failure categories for overfitting or reward hacking. \\
\bottomrule
\end{tabularx}
}
\end{minipage}
\end{center}

\section{Detailed evaluation agenda and limitations}
\label{app:evaluation-agenda}

\paragraph{Purpose of this appendix.}
This appendix expands the evaluation protocol in Table~\ref{tab:decisive}. The goal is not to require every TTT paper to run every possible memory test. Rather, the evaluation should match the claim. A paper claiming stream adaptation can headline stream loss, future-token loss, long-context accuracy, throughput, or reward. A paper claiming deployment-time memory, personalization, correction, or agent memory should additionally report later behavior under a deployment-like update episode after the original support context is unavailable.

\paragraph{Diagnostic role of the pilot.}
The diagnostic in Section~\ref{sec:controlled-diagnostic} partially follows this agenda. It specifies exact-context controls, reports BM25-style retrieval and replacement-memory checks, separates answer-likelihood and generated-answer reporting, discloses update-time and token-budget details, and adds a same-family Qwen3 scale axis, a 1.7B three-seed replication, a stress update, mutually exclusive conflict scoring, and small preference/correction and procedure/action probes. Full benchmark construction and representative method comparisons are left for future work.

\begin{takeawaybox}
\textbf{Diagnostic role:} The LoRA/Qwen pilot isolates one evidential question: whether support-loss and answer-likelihood gains are sufficient for generated deployment behavior. The result shows that these proxy and bridge metrics can improve without free-form recall, motivating separate reporting of mechanism-level and behavioral evidence.
\end{takeawaybox}

\paragraph{Minimal D-level evidence by claim type.}
D-level evidence is claim-specific. The required behavioral bundle changes with the claim, but in every case the learned information should affect later behavior after the original support context is removed. Table~\ref{tab:modular-d-evidence} gives minimum evidence shapes for common deployment-memory claims.

\begin{center}
\begin{minipage}{\linewidth}
\captionof{table}{Minimal D-level evidence by deployment claim type. The required bundle changes with the claim, but in every case the later behavior must match the claimed use case after the original support context is removed.}
\label{tab:modular-d-evidence}
\vspace{3pt}

\centering
{\tiny
\setlength{\tabcolsep}{1.9pt}
\renewcommand{\arraystretch}{1.08}
\begin{tabularx}{\linewidth}{@{}L{0.16\linewidth} Y Y L{0.20\linewidth}@{}}
\toprule
\tabhead{Claim type} & \tabhead{Minimal D-level behavioral evidence} & \tabhead{Required failure categories} & \tabhead{Matched baseline} \\
\midrule
Factual user memory & no-support-context answer; direct and paraphrased query; delayed query & correct / wrong / abstain; locality & exact context; retrieval memory \\
\addlinespace[1pt]
Preference personalization & later choice or response-style change under plausible alternatives & follows preference / ignores / overgeneralizes / unsafe conflict & profile prompt; retrieval profile \\
\addlinespace[1pt]
Correction / overwrite & corrected-only response after stale fact is contradicted & corrected-only / stale-only / both / neither & replacement memory with stale item removed \\
\addlinespace[1pt]
Procedure learning & held-out task where the learned rule changes action, not just string recall & correct transfer / memorized surface form / unrelated degradation & explicit instruction; tool memory \\
\addlinespace[1pt]
Agent/tool memory & learned fact changes tool call, route, or argument & correct action / wrong action / leakage / stale action & memory-augmented agent \\
\addlinespace[1pt]
Reward-backed discovery & later held-out reward or task gain under the same discovery budget & solved / overfit / reward hacking / unrelated damage & no-update; search or retrieval baseline \\
\bottomrule
\end{tabularx}
}
\end{minipage}
\end{center}

\paragraph{Operational behavioral templates.}
The same claim-specific idea can be made more operational by specifying the support episode, the no-context query, paraphrase or delay condition, locality or conflict check, and matched baseline. Table~\ref{tab:claim-specific-templates} gives templates for common deployment-memory settings. These are intended as minimal behavioral shapes, not as a universal benchmark.

\begin{center}
\begin{minipage}{\linewidth}
\captionof{table}{Claim-specific behavioral templates.}
\label{tab:claim-specific-templates}
\vspace{3pt}

\centering
{\tiny
\setlength{\tabcolsep}{1.6pt}
\renewcommand{\arraystretch}{1.07}
\begin{tabularx}{\linewidth}{L{1.55cm} Y Y Y Y L{1.85cm}}
\toprule
\tabhead{Claim} & \tabhead{Support episode} & \tabhead{No-context query} & \tabhead{Paraphrase / delay} & \tabhead{Locality / conflict} & \tabhead{Matched baseline} \\
\midrule
Factual memory & One sparse user fact absent from the pre-update prompt. & Ask for the fact after removing support context. & Direct and semantic paraphrase; at least one intervening task or session break. & Unrelated facts or abilities should remain stable. & Exact context; retrieval memory. \\
\addlinespace[1pt]
Preference personalization & User states a stable preference under plausible alternatives. & Later choice or response style should reflect the preference. & Query with different surface form and scenario. & Check overgeneralization, unsafe preference conflicts, and unrelated preferences. & Profile prompt; retrieved profile. \\
\addlinespace[1pt]
Correction / overwrite & A stale fact is explicitly contradicted by a current correction. & Ask for the current answer without showing either support item. & Query correction with rewording or after unrelated turns. & Score corrected-only, stale-only, both, and neither. & Replacement memory with stale item removed. \\
\addlinespace[1pt]
Procedure learning & User gives a rule or procedure that changes task execution. & Held-out task requires applying the learned rule. & Rephrase the procedure or change surface variables. & Check unrelated procedures and mistaken transfer. & Explicit instruction; tool memory. \\
\addlinespace[1pt]
Agent/tool memory & User fact or rule should affect routing, tool choice, or arguments. & Later task requires the remembered item for action selection. & Change wording and separate the action from the support turn. & Score wrong tool, stale action, leakage, and unrelated route changes. & Memory-augmented agent. \\
\bottomrule
\end{tabularx}
}
\end{minipage}
\end{center}

\paragraph{Direct and paraphrased retrieval.}
A deployment-time update should be tested on direct questions and paraphrases that require extracting the newly introduced fact, preference, correction, or procedure after the support context is removed. Teacher-forced answer likelihood and free generation should be reported separately, because the gap between them is itself informative. A system can move probability mass toward the correct answer under teacher forcing while still failing to produce the answer under open-ended generation.

\paragraph{Retention, locality, and conflict.}
Updated knowledge should be queried after unrelated intervening turns or tasks. Evaluations should also include locality probes and conflict-resolution tests, because a useful update should not indiscriminately distort unrelated behavior or preserve stale information after a correction. If the claim concerns procedures or agents, the query should require the updated information to select a rule, route, tool, or action rather than merely repeat a string.

\paragraph{Matched explicit-memory baselines.}
If the intended use case is a personalized assistant or agent, prompt accumulation, long-context inference, retrieval-augmented memory, and other non-parametric memory systems are natural baselines. These baselines should use matched evidence and report context cost, latency, update budget, and failure categories, because a parametric update is most compelling when it improves behavior under constraints where explicit memory is costly or unavailable. Stream adaptation should also be separated from one-shot write-in through reset-vs-stream controls, since cumulative gains across a stream provide different evidence from learning from a single support item.

\begin{center}
\begin{minipage}{\linewidth}
\captionof{table}{Matched explicit-memory baseline protocol.}
\label{tab:matched-baseline-protocol}
\vspace{3pt}

\centering
{\footnotesize
\setlength{\tabcolsep}{3pt}
\renewcommand{\arraystretch}{1.08}
\begin{tabularx}{\linewidth}{L{2.25cm} Y Y}
\toprule
\tabhead{Baseline} & \tabhead{What it controls} & \tabhead{Matching requirement} \\
\midrule
No update & Whether the update helps relative to the base model or unmodified system. & Same query set, generation budget, scoring rule, and random seed policy when applicable. \\
\addlinespace[1pt]
Exact context & Whether the support evidence is sufficient when directly visible. & Same support item, query, scoring rule, and generation budget as the parametric-update condition. \\
\addlinespace[1pt]
Long-context prompting & Whether keeping the support evidence in context solves the task. & Same support information, same downstream query, disclosed context length, and comparable generation budget. \\
\addlinespace[1pt]
Retrieval memory & Whether the evidence can be stored outside weights and retrieved for the same task. & Same memory item, query set, answer scoring, retrieval index disclosure, and context/latency budget. \\
\addlinespace[1pt]
Replacement memory & Whether a correction can remove stale information before answering. & Same stale/current evidence, mutually exclusive corrected/stale/both/neither categories, and current-only memory state. \\
\addlinespace[1pt]
Memory-augmented agent & Whether learned information changes a tool call, route, or argument through explicit state. & Same support episode, tool set, action space, cost budget, and leakage/stale-action scoring. \\
\bottomrule
\end{tabularx}
}
\end{minipage}
\end{center}

In this paper, the easy BM25-style factual condition is a usability control: it verifies that the sparse support facts are answerable when explicit memory retrieves the intended sentence. The harder paraphrase and stale/current retrieval checks are failure-mode probes, showing why stronger explicit-memory baselines should include recency handling, semantic retrieval, reranking, or conflict resolution rather than naive lexical matching alone.

\paragraph{Evidence tiers for deployment-memory language.}
The strength of the claim should track the strength of the behavioral evidence. Table~\ref{tab:evidence-tiers} gives three rough tiers. The tiers are not meant to define a benchmark leaderboard; they are a wording guide for authors and evaluators.

\begin{center}
\begin{minipage}{0.95\linewidth}
\captionof{table}{Evidence tiers for deployment-memory language.}
\label{tab:evidence-tiers}
\vspace{3pt}

\centering
{\footnotesize
\setlength{\tabcolsep}{3pt}
\renewcommand{\arraystretch}{1.08}
\begin{tabularx}{\linewidth}{L{2.45cm} Y}
\toprule
\tabhead{Tier} & \tabhead{Appropriate evidence and wording} \\
\midrule
Minimum acceptable & A small no-support-context behavioral probe matched to the claim, with direct/paraphrased queries, disclosed scoring, and a no-update or exact-context control. Supports narrow deployment-memory wording under the tested setting. \\
\addlinespace[1pt]
Strong evidence & Adds delay, locality, conflict categories when relevant, multiple seeds or model sizes, and matched explicit-memory baselines. Supports robust claim wording within the tested update budget. \\
\addlinespace[1pt]
Deployment-grade & Adds realistic multi-session data, deletion/forgetting or user-isolation tests when personal data is stored, latency/privacy/context-cost accounting, and stronger external-memory comparators. Supports deployment-oriented system claims. \\
\bottomrule
\end{tabularx}
}
\end{minipage}
\end{center}

\paragraph{Wording examples.}
The decision protocol is meant to be operational. Table~\ref{tab:claim-wording-examples} gives examples of wording that stays within the evidence level and wording that calls for D-level behavioral evidence.

\begin{center}
\begin{minipage}{0.92\linewidth}
\captionof{table}{Claim wording examples.}
\label{tab:claim-wording-examples}
\vspace{3pt}

\centering
{\footnotesize
\setlength{\tabcolsep}{4pt}
\renewcommand{\arraystretch}{1.08}
\begin{tabularx}{\linewidth}{Y Y}
\toprule
\tabhead{Better wording} & \tabhead{Too strong unless D-level evidence exists} \\
\midrule
Improves future-token prediction after online updates & Learns user preferences at deployment time \\
\addlinespace[1pt]
Improves domain or task performance under the stated objective & Performs post-deployment continual learning \\
\addlinespace[1pt]
Compresses dense context into a test-time state & Forms persistent assistant memory \\
\addlinespace[1pt]
Improves answer likelihood or candidate rank for support facts & Can recall newly learned facts in deployment \\
\addlinespace[1pt]
Shows bridge evidence for internalization under this protocol & Enables self-updating assistant behavior \\
\bottomrule
\end{tabularx}
}
\end{minipage}
\end{center}

\paragraph{Claim discipline.}
Papers should title and frame their contributions at the level their evidence supports. Continuous-stream adaptation, target-domain TTL, formal test-time discovery, context compression, and deployment-time learning are all legitimate targets. Evidence from one regime should not inherit the behavioral expectations of another. In particular, proxy and bridge metrics should remain in the report, but they should be described as mechanism-level or intermediate evidence unless the later behavior required by the claim is directly tested.

\paragraph{Additional limitations.}
This diagnostic pass adds a same-family Qwen3 size axis, hard-negative retrieval controls, a 1.7B three-seed replication, conflict scoring, and small preference/correction and procedure/action probes. It leaves several directions open: cross-family model comparisons, a second parametric update mechanism, representative TTT/TTL systems, richer procedure-following tasks, and multi-seed runs for every model size. The confidence intervals are across examples unless explicitly marked as seed replication. Reward-rich settings such as math, code, kernels, and scientific optimization remain an important frontier. These limitations reinforce the central recommendation: specify the evaluation target, disclose the scoring rule, report failure categories, and avoid compressing all evidence into one loss number.

\section{Paper-level audit sheet}
\label{app:audit-sheet}

This appendix expands the claim-calibrated audit summarized in the main text. The level column records the strongest claim directly supported by the reported evidence as summarized in the paper, while the last column records the shortest behavioral test that would be needed before using the result as D-level deployment-time learning evidence.

\paragraph{Audit protocol.}
We screened papers available through April 2026 from the paper's bibliography plus targeted searches using terms such as \emph{LLM test-time training}, \emph{test-time learning}, \emph{long-context TTT}, \emph{context memory}, \emph{self-adapting language models}, \emph{parametric memory}, \emph{personalization}, \emph{continual learning}, and \emph{assistant memory}. A paper enters the audit when its title, abstract, introduction, or motivation explicitly invokes test-time learning/training, memory, persistence, context absorption, self-improvement, personalization, continual learning, or assistant memory. We screened 41 candidate records: 24 are coded below, 5 retrieval-only or external-memory systems are retained as background baselines, 9 model-editing/safety/background papers are cited for context but excluded from claim-level coding, and 3 non-LLM or architecture-adjacent papers are excluded from the audit table. This is single-author coding with sensitivity recoding, intended as a claim-calibration audit rather than a systematic review, prevalence estimate, or inter-coder reliability study. To make disagreements inspectable, we provide paper-level source phrases, boundary-case coding rationales, and sensitivity effects rather than asking readers to trust aggregate counts.

S-level coding means the reported evidence directly supports stream/domain/task adaptation; B-level coding means it supports a bridge mechanism such as internalization, compression, parametric memory, or self-adaptation; D-level coding means it directly tests sparse deployment information after paraphrase, delay, conflict, locality, or action use. Borderline cases are coded to the strongest level directly supported by the reported evaluation, not by the broadest motivation sentence. Eight papers are treated as boundary cases for sensitivity: TTT layers, LaCT, Not Just Context, TTT-E2E, PERK, Locas, MEMORYLLM, and Absorber LLM. Reclassifying these S/B cases changes the counts without changing the paper's decision rule, because none supplies the full D-level behavioral bundle for sparse user deployment claims.

\begin{center}
\begin{minipage}{0.95\linewidth}
\captionof{table}{Audit field summary.}
\label{tab:audit-field-summary}
\vspace{3pt}

\centering
{\footnotesize
\setlength{\tabcolsep}{3pt}
\renewcommand{\arraystretch}{1.08}
\begin{tabularx}{\linewidth}{L{2.45cm} Y}
\toprule
\tabhead{Field} & \tabhead{Definition} \\
\midrule
Search source & Bibliography seeding plus targeted searches through April 2026. \\
\addlinespace[1pt]
Search terms & LLM test-time training, test-time learning, long-context TTT, context memory, self-adapting language models, parametric memory, personalization, continual learning, and assistant memory. \\
\addlinespace[1pt]
Inclusion rule & Title, abstract, introduction, or motivation invokes TTT/TTL, memory, persistence, context absorption, self-improvement, personalization, continual learning, or assistant memory. \\
\addlinespace[1pt]
Exclusion rule & Retrieval-only/external-memory systems, model-editing and safety background, non-LLM TTT, and architecture-adjacent memory papers are retained as context but not coded as LLM TTT evidence. \\
\addlinespace[1pt]
Coding target & Strongest claim level directly supported by reported evaluations, not by the broadest motivation sentence. \\
\bottomrule
\end{tabularx}
}
\end{minipage}
\end{center}

\begin{center}
\begin{minipage}{0.82\linewidth}
\captionof{table}{Candidate accounting.}
\label{tab:candidate-accounting}
\vspace{3pt}

\centering
{\footnotesize
\setlength{\tabcolsep}{4pt}
\renewcommand{\arraystretch}{1.08}
\begin{tabularx}{\linewidth}{L{4.2cm} c Y}
\toprule
\tabhead{Category} & \tabhead{Count} & \tabhead{Treatment} \\
\midrule
Coded stream, long-context, TTL, or reward-rich task adaptation & 10 & Claim-level coded as S or S/B. \\
\addlinespace[1pt]
Coded bridge/internalization or parametric-memory evidence & 8 & Claim-level coded as B. \\
\addlinespace[1pt]
Coded D-level assistant-memory target benchmarks & 6 & Claim-level coded as D target behavior. \\
\addlinespace[1pt]
External-memory baselines & 5 & Retained as matched-baseline context, not LLM TTT claim-coded. \\
\addlinespace[1pt]
Model-editing and safety background & 9 & Retained for evaluation lessons and risk framing. \\
\addlinespace[1pt]
Non-LLM or architecture-adjacent exclusions & 3 & Retained only as boundary context. \\
\midrule
Total screened & 41 & 24 coded; 17 background or excluded from claim-level coding. \\
\bottomrule
\end{tabularx}
}
\end{minipage}
\end{center}

\paragraph{Boundary-case coding worksheet.}
We provide the following worksheet for inspecting the eight boundary cases used in the audit sensitivity discussion.

\begin{center}
\begin{minipage}{\linewidth}
\captionof{table}{Boundary-case coding worksheet.}
\label{tab:boundary-case-coding}
\vspace{3pt}

\centering
{\tiny
\setlength{\tabcolsep}{1.7pt}
\renewcommand{\arraystretch}{1.05}
\begin{tabularx}{\linewidth}{L{1.75cm} L{1.0cm} Y Y}
\toprule
\tabhead{Paper} & \tabhead{Level} & \tabhead{Boundary reason} & \tabhead{Sensitivity effect} \\
\midrule
TTT layers & S/B & Fast weights improve sequence and long-context behavior; sparse delayed deployment memory remains untested & Recoding as S or B changes counts only; D-level still unsupported. \\
\addlinespace[1pt]
LaCT & S/B & Large-chunk TTT improves efficiency and long-context performance; D-level sparse behavior remains to be tested & Recoding as S or B leaves required sparse behavior unchanged. \\
\addlinespace[1pt]
Not just context & S/B & Context-specific updates are close to bridge evidence; matched delayed deployment behavior is missing & Recoding as S or B does not license personalization language. \\
\addlinespace[1pt]
TTT-E2E & B & Long-context compression and needle-style recall provide bridge evidence for deployment-memory evaluation & Treating as S/B would still require sparse semantic write-in. \\
\addlinespace[1pt]
PERK & B & Parameter-efficient test-time learning suggests internalization; sparse preference/correction behavior is missing & Boundary remains below D until user-facing behavior is tested. \\
\addlinespace[1pt]
Locas & B & Parametric memory framing gives bridge evidence; D-level conflict/locality behavior remains incomplete & Stronger memory framing raises, rather than removes, the D-level burden. \\
\addlinespace[1pt]
MEMORYLLM & B & Persistent memory pools support memory-like behavior and call for matched parametric TTT/no-context ablation for TTT claims & Counts could move toward D target, but ordinary parametric TTT claims still need ablation. \\
\addlinespace[1pt]
Absorber LLM & B & Context absorption and synchronization are bridge evidence; sparse user write-in with behavioral ablation is missing & Recoding as B or S/B leaves no-context deployment behavior untested. \\
\bottomrule
\end{tabularx}
}
\end{minipage}
\end{center}

\paragraph{Background and exclusion records.}
The following records were screened because they are useful comparators, cautionary evaluation background, or boundary cases, but they are not coded as direct LLM TTT evidence in Tables~\ref{tab:audit-stream-long-context}--\ref{tab:audit-d-targets}.

\begin{center}
\begin{minipage}{\linewidth}
\captionof{table}{Background and exclusion records.}
\label{tab:background-exclusion-records}
\vspace{3pt}

\centering
{\tiny
\setlength{\tabcolsep}{2pt}
\renewcommand{\arraystretch}{1.05}
\begin{tabularx}{\linewidth}{L{2.75cm} L{1.75cm} Y Y}
\toprule
\tabhead{Record} & \tabhead{Role} & \tabhead{Evidence retained} & \tabhead{Reason not claim-coded as LLM TTT} \\
\midrule
MemoryBank \citep{zhong2024memorybank} & External memory & Long-term memory storage and retrieval. & Baseline context, not parametric TTT evidence. \\
\addlinespace[1pt]
LongMem \citep{wang2023longmem} & External memory & Retrieval/cache memory for long context. & Baseline context, not TTT update evidence. \\
\addlinespace[1pt]
MemGPT \citep{packer2023memgpt} & External memory & Paged memory and context management. & Baseline context, not parametric TTT evidence. \\
\addlinespace[1pt]
Mem0 \citep{chhikara2025mem0} & External memory & Scalable explicit memory system. & Baseline context for production memory. \\
\addlinespace[1pt]
Dynamic Cheatsheet \citep{suzgun2026dynamiccheatsheet} & External memory & Adaptive memory for reusable problem-solving snippets. & Explicit-memory comparator rather than ordinary weight update. \\
\addlinespace[1pt]
ROME \citep{meng2022rome} & Model editing & Efficacy, generalization, and specificity framing. & Background caution for evaluating state changes. \\
\addlinespace[1pt]
MEMIT \citep{meng2023memit} & Model editing & Mass editing and specificity. & Background caution for interference and scale. \\
\addlinespace[1pt]
Edit-failure benchmark \citep{hoelscherobermaier2023detecting} & Model editing & Specificity failures after edits. & Background evaluation lesson. \\
\addlinespace[1pt]
MQuAKE \citep{zhong2023mquake} & Model editing & Multi-hop editing evaluation. & Background evaluation lesson. \\
\addlinespace[1pt]
Long-form editing evaluation \citep{rosati2024leme} & Model editing & Long-form behavior after edits. & Background evaluation lesson. \\
\addlinespace[1pt]
Editing at scale \citep{gupta2024editing-scale} & Model editing & Forgetting under many edits. & Background evaluation lesson. \\
\addlinespace[1pt]
Should we edit LMs? \citep{li2024shouldedit} & Model editing & Edit-necessity and evaluation critique. & Background evaluation lesson. \\
\addlinespace[1pt]
Mirage of model editing \citep{yang2025mirage} & Model editing & In-the-wild edit evaluation failures. & Background evaluation lesson. \\
\addlinespace[1pt]
TTT safety guardrails \citep{antonelli2026safetyguardrails} & Safety background & Safety impact of TTT. & Risk context rather than memory-claim evidence. \\
\addlinespace[1pt]
Original TTT \citep{sun2020ttt} & Historical boundary & Self-supervised TTT for distribution shifts. & Non-LLM deployment-memory setting. \\
\addlinespace[1pt]
Titans \citep{behrouz2025titans} & Architecture-adjacent & Memory architecture for test-time memorization. & Boundary context, not LLM TTT claim coding. \\
\addlinespace[1pt]
ATLAS \citep{behrouz2025atlas} & Architecture-adjacent & Context memorization architecture. & Boundary context, not LLM TTT claim coding. \\
\bottomrule
\end{tabularx}
}
\end{minipage}
\end{center}

\paragraph{Worked examples.}
\emph{TTT-NN:} S pass, D no. It updates on nearest-neighbor text and reports local language-model improvement, so it supports adaptation to nearby retrieved evidence; deployment-memory use would require delayed no-context recall, locality, and matched memory baselines. \emph{PERK/Locas:} B pass, D incomplete. They explicitly move toward parameter-efficient or locally supported memories, so their evidence is closer to internalization; D-level assistant-memory claims still need sparse user facts, preferences, and corrections under paraphrase, delay, conflict, and matched external-memory baselines. \emph{LongMemEval/MemoryBench:} D target behavior yes. They evaluate multi-session or service-time memory behavior, so they instantiate the target capability; a TTT paper borrowing that motivation should report comparable behavior under a stated update budget. \emph{This pilot:} proxy and bridge evidence yes, D behavior no for the tested one-step LoRA setup; matched retrieval and replacement-memory checks are reported, so the result supports the proxy-insufficiency point: proxy and bridge improvements should not be treated as deployment-memory evidence without matching behavioral success.

\begin{center}
\begin{minipage}{\linewidth}
\captionof{table}{Paper-level audit for stream and long-context TTT evidence.}
\label{tab:audit-stream-long-context}
\vspace{3pt}

\centering
{\tiny
\setlength{\tabcolsep}{2.0pt}
\renewcommand{\arraystretch}{1.04}
\begin{tabularx}{\linewidth}{L{2.25cm} L{2.25cm} L{0.55cm} L{3.2cm} Y}
\toprule
\tabhead{Paper} & \tabhead{Source phrase} & \tabhead{Level} & \tabhead{Coding rationale} & \tabhead{Minimal evidence before D} \\
\midrule
Dynamic evaluation \citep{krause2018dynamic} & ``Dynamic evaluation'' & S & Updates on recent sequence context and evaluates sequence-model loss. & Delayed no-context recall and locality. \\
\addlinespace[1pt]
Revisiting dynamic evaluation \citep{rannentriki2024dynamic} & ``Online adaptation'' & S & Studies online adaptation for LLMs with stream-style evidence. & Sparse user fact write-in after delay. \\
\addlinespace[1pt]
TTT-NN \citep{hardt2024tttnn} & ``Test-Time Training on Nearest Neighbors'' & S & Adapts from retrieved neighbors and evaluates local LM improvement. & Paraphrased retrieval without neighbor context. \\
\addlinespace[1pt]
TTT layers \citep{sun2024tttlayers} & ``Learn at Test Time'' & S/B & Fast-weight hidden states improve sequence modeling and long-context behavior. & Reset-vs-stream plus sparse delayed behavior. \\
\addlinespace[1pt]
LaCT \citep{zhang2025lact} & ``Test-Time Training Done Right'' & S/B & Emphasizes efficient large-chunk TTT and long-context performance. & One-shot fact/preference retention and conflict. \\
\addlinespace[1pt]
In-Place TTT \citep{feng2026inplace} & ``In-Place Test-Time Training'' & S & Aligns online updates with next-token prediction objectives. & Behavioral recall after removing support context. \\
\addlinespace[1pt]
Not just context \citep{bansal2025notjustcontext} & ``things in Context'' & S/B & Tests context-specific updates for long-context LLMs. & Matched retrieval baseline plus delayed behavior. \\
\addlinespace[1pt]
TTT-E2E \citep{tandon2025e2e} & ``End-to-End Test-Time Training for Long Context'' & B & Targets long-context compression and needle-style recall. & Sparse semantic write-in under paraphrase and delay. \\
\bottomrule
\end{tabularx}
}
\end{minipage}
\end{center}

\begin{center}
\begin{minipage}{\linewidth}
\captionof{table}{Paper-level audit for bridge evidence toward memory-like claims.}
\label{tab:audit-bridge-evidence}
\vspace{3pt}

\centering
{\tiny
\setlength{\tabcolsep}{2.0pt}
\renewcommand{\arraystretch}{1.04}
\begin{tabularx}{\linewidth}{L{2.25cm} L{2.25cm} L{0.55cm} L{3.2cm} Y}
\toprule
\tabhead{Paper} & \tabhead{Source phrase} & \tabhead{Level} & \tabhead{Coding rationale} & \tabhead{Minimal evidence before D} \\
\midrule
PERK \citep{chen2026perk} & ``Parameter-Efficient Test-Time Learning'' & B & Uses parameter-efficient updates for long-context reasoning. & Sparse facts/preferences tested after delay. \\
\addlinespace[1pt]
Locas \citep{lu2026locas} & ``Parametric Memories'' & B & Frames models as initializers for locally supported memories. & Conflict overwrite and locality under user corrections. \\
\addlinespace[1pt]
MEMORYLLM \citep{wang2024memoryllm} & ``Self-Updatable'' & B & Implements persistent memory pools for self-updating behavior. & Matched parametric TTT and no-context ablation. \\
\addlinespace[1pt]
TLM \citep{hu2025tlm} & ``Test-Time Learning'' & S & Optimizes input likelihood at test time. & User-specific write-in and generated recall. \\
\addlinespace[1pt]
SyTTA \citep{xu2026sytta} & ``Test-Time Adaptation'' & S & Uses label-free adaptation signals rather than new user memory. & Delayed paraphrase and locality tests. \\
\addlinespace[1pt]
TTRL \citep{zuo2025ttrl} & ``Test-Time Reinforcement Learning'' & S & Improves tasks with reward-style test-time updates. & Reward-poor user facts and corrections. \\
\addlinespace[1pt]
SEAL \citep{zweiger2025seal} & ``Self-Adapting Language Models'' & B & Generates update data and reports self-adaptation gains. & Novel-information recall after context removal. \\
\addlinespace[1pt]
TT-SI \citep{acikgoz2025ttsi} & ``Self-Improving LLM Agents'' & B & Studies self-improvement for agents at test time. & Personalization, conflict, and retention probes. \\
\addlinespace[1pt]
TTT-Discover \citep{yuksekgonul2026tttdiscover} & ``Discover at Test Time'' & B & Uses verifiable rewards for active test-time discovery. & Weak-evidence assistant memory without rewards. \\
\addlinespace[1pt]
Absorber LLM \citep{zhang2026absorber} & ``Test-Time Training'' & B & Studies causal synchronization and context absorption. & Sparse user write-in with causal ablation. \\
\bottomrule
\end{tabularx}
}
\end{minipage}
\end{center}

\begin{center}
\begin{minipage}{\linewidth}
\captionof{table}{Paper-level audit for D-level assistant-memory targets.}
\label{tab:audit-d-targets}
\vspace{3pt}

\centering
{\tiny
\setlength{\tabcolsep}{2.0pt}
\renewcommand{\arraystretch}{1.04}
\begin{tabularx}{\linewidth}{L{2.25cm} L{2.25cm} L{0.55cm} L{3.2cm} Y}
\toprule
\tabhead{Paper} & \tabhead{Source phrase} & \tabhead{Level} & \tabhead{Coding rationale} & \tabhead{Remaining gap for TTT claims} \\
\midrule
LoCoMo \citep{maharana2024locomo} & ``Very Long-Term Conversational Memory'' & D & Tests long-horizon conversational memory behavior. & Add matched parametric-update comparison. \\
\addlinespace[1pt]
LongMemEval \citep{wu2024longmemeval} & ``Long-Term Interactive Memory'' & D & Evaluates memory in chat-assistant interactions. & Add TTT update budget and ablations. \\
\addlinespace[1pt]
MemoryAgentBench \citep{hu2025memoryagentbench} & ``Incremental Multi-Turn Interactions'' & D & Tests memory through incremental agent interactions. & Compare against prompt/retrieval and parametric TTT. \\
\addlinespace[1pt]
MemoryBench \citep{ai2025memorybench} & ``Memory and Continual Learning'' & D & Targets service-time memory and continual learning. & Add explicit TTT method comparison if claimed. \\
\addlinespace[1pt]
MemoryCD \citep{zhang2026memorycd} & ``Lifelong Cross-Domain Personalization'' & D & Tests long-context user memory for personalization. & Separate explicit memory from parametric update. \\
\addlinespace[1pt]
Mem2ActBench \citep{shen2026mem2actbench} & ``Long-Term Memory Utilization'' & D & Measures memory use in downstream agent actions. & Add matched TTT write-in and action ablation. \\
\bottomrule
\end{tabularx}
}
\end{minipage}
\end{center}

\section{Diagnostic robustness checks}
\label{app:diagnostic-robustness}

\paragraph{Pilot experiment details.}
Unless otherwise noted, diagnostic runs use H100 GPUs, seed 13, LoRA rank 8, alpha 16, dropout 0.0, learning rate $5 \times 10^{-4}$, and a 24-token generation budget. The factual probes use 48 prompts per prompt type, the overwrite probes use 24 conflicts, and the locality probes use 144 unrelated prompts. We score generated answers as correct when the normalized first answer line contains the normalized target answer. We define $\Delta\mathrm{NLL}=\mathrm{NLL}_{\mathrm{after}}-\mathrm{NLL}_{\mathrm{before}}$, so negative deltas indicate improved teacher-forced loss after the update.

The matched external-memory baselines receive the same support evidence as the LoRA update. The \emph{exact-context} baseline prepends the support sentence directly to the evaluation prompt. The \emph{BM25-style retrieval} baseline builds a fixed memory bank whose retrieval unit is the support sentence. For each true fact, we add three hard negatives: the same object with a different prefix, the same prefix with a different object, and an answer-format distractor. We report both top-1 retrieval hit, counted as correct when the retrieved fact id matches the target id, and end-to-end answer success after prompting with \texttt{Retrieved memory: <support>} followed by the original query. For conflicts, the \emph{replacement-memory} baseline discards the stale support, keeps only the correction sentence, and then answers the current-code query.

The supporting checks are deliberately narrow. On 1.7B only, we run a continuous-text proxy-regime check by splitting 32 passages into \emph{support}, \emph{future\_1}, and \emph{future\_2}, updating on the support chunk, and comparing one-step, stream, and reset evaluation. We rerun the 1.7B factual/conflict core probes with two additional seeds. We also add small 24-item 1.7B preference/correction and procedure/action mini-tasks to test whether the proxy/behavior pattern is limited to nonce access-code strings.

\begin{center}
\begin{minipage}{\linewidth}
\captionof{table}{Robustness and stress checks for the 1.7B diagnostic setting.}
\label{tab:robustness-stress-checks}
\vspace{3pt}

\centering
{\footnotesize
\setlength{\tabcolsep}{4pt}
\renewcommand{\arraystretch}{1.08}
\begin{tabularx}{\linewidth}{L{2.15cm} Y}
\toprule
\tabhead{Setting} & \tabhead{Key result} \\
\midrule
Seeds 13 / 21 / 42 &
For the 1.7B model, the core pattern is unchanged across seeds: greedy direct/paraphrased/delayed recall remains 0.0\% ($0/48$), BM25 hit remains 100.0\% ($48/48$), memory answer remains 100.0\% ($48/48$), LoRA conflict correction remains 0.0\% ($0/24$), and locality ranges from 95.1\%--97.9\% ($137$--$141/144$). \\
\addlinespace[2pt]
Preference/correction mini-task &
Preference updates improve proxy loss, with $\Delta\mathrm{NLL}(\mathrm{S}/\mathrm{D}/\mathrm{P}) = -1.524/-0.709/-0.629$, while greedy direct/paraphrased/delayed behavior remains 0.0\% ($0/24$). Exact-context answers are 87.5\%--91.7\% ($21$--$22/24$), and replacement memory is corrected-only for most preference corrections. \\
\addlinespace[2pt]
Procedure mini-task &
Procedure updates also improve proxy loss, with $\Delta\mathrm{NLL}(\mathrm{S}/\mathrm{D}/\mathrm{P}) = -1.219/-0.854/-0.785$, while greedy direct/paraphrased/delayed behavior remains 0.0\% ($0/24$). Exact-context answers are 58.3\% ($14/24$), 33.3\% ($8/24$), and 16.7\% ($4/24$) for the three procedure prompt types, and replacement memory resolves 87.5\% ($21/24$) of conflicts. \\
\addlinespace[2pt]
Update-strength sweep &
We use 1.7B LoRA sweep to cover rank-8 steps 1/4/16 and rank-32 steps 1/4. Rank-8 steps 1/4 have 0/48 direct/paraphrased/delayed recall with 141/144 and 135/144 locality; the rank-8 16-step stress point has 35/48 direct, 26/48 paraphrased, and 35/48 delayed recall with 14/144 locality. Rank-32 steps 1/4 also have 0/48 recall, with 143/144 and 131/144 locality. \\
\addlinespace[2pt]
Conflict categories &
Mutually exclusive conflict scoring shows that one-step LoRA is neither-corrected-nor-stale in all 72 conflicts across Qwen3 1.7B/4B/8B. Replacement memory is corrected-only in 68/72 cases and both corrected-and-stale in 4/72 cases, all from the 1.7B run. \\
\addlinespace[2pt]
Continuous proxy check &
For 1.7B, one-step support/F1/F2 deltas are $-0.636{\pm}0.042$, $-0.063{\pm}0.012$, and $-0.058{\pm}0.010$, with stream F1 at $-0.384{\pm}0.059$. \\
\addlinespace[2pt]
16-step stress update &
As discussed in the main text, recall rises to 72.9\% ($35/48$) on direct and delayed prompts and 54.2\% ($26/48$) on paraphrased prompts, while locality drops sharply to 9.7\% ($14/144$). We treat this setting as a locality stress test, not as a practical operating point. \\
\bottomrule
\end{tabularx}
}
\end{minipage}
\end{center}

\paragraph{Conflict-overwrite scoring.}
Correction claims require mutually exclusive categories. A response counts as successful only when it contains the corrected answer and excludes the stale answer. Responses containing both corrected and stale information are failures, even if a non-exclusive scorer would mark the corrected answer as present.

\begin{center}
\begin{minipage}{0.88\linewidth}
\captionof{table}{Mutually exclusive conflict-overwrite categories. Counts are out of 24 conflicts per model. Corrected-only is the strict D-level success criterion; both corrected and stale is counted as failure.}
\label{tab:conflict-overwrite-categories}
\vspace{3pt}

\centering
{\scriptsize
\setlength{\tabcolsep}{4.0pt}
\renewcommand{\arraystretch}{1.08}
\begin{tabular}{@{}llcccc@{}}
\toprule
Model & Method & Corrected-only & Both & Neither & Locality preserved \\
\midrule
Qwen3-1.7B & LoRA update & $0$ & $0$ & $24$ & $68/72$ \\
Qwen3-1.7B & Replacement memory & $20$ & $4$ & $0$ & -- \\
Qwen3-4B & LoRA update & $0$ & $0$ & $24$ & $71/72$ \\
Qwen3-4B & Replacement memory & $24$ & $0$ & $0$ & -- \\
Qwen3-8B & LoRA update & $0$ & $0$ & $24$ & $72/72$ \\
Qwen3-8B & Replacement memory & $24$ & $0$ & $0$ & -- \\
\bottomrule
\end{tabular}
}
\end{minipage}
\end{center}

The one-step LoRA update produces neither code in all conflict cases, indicating overwrite non-recall rather than successful correction. This is a failure of behavioral access, not merely a failure of conflict arbitration. Replacement memory is usually corrected-only, although the 1.7B model sometimes includes both corrected and stale codes in the same response.

\paragraph{Stress update tradeoff.}
A stronger update can produce generated recall, but recall alone is not sufficient for deployment-memory claims. Locality probes ask unrelated questions whose answers should remain unchanged after the update.

\begin{center}
\begin{minipage}{0.82\linewidth}
\captionof{table}{Stress update tradeoff in the Qwen3-1.7B factual setting.}
\label{tab:stress-update-tradeoff}
\vspace{3pt}

\centering
{\scriptsize
\setlength{\tabcolsep}{4.4pt}
\renewcommand{\arraystretch}{1.08}
\begin{tabular}{@{}lcccc@{}}
\toprule
\textbf{Setting} & \textbf{Direct Recall $\uparrow$} & \textbf{Paraphrased Recall $\uparrow$} & \textbf{Delayed Recall $\uparrow$} & \textbf{Locality $\uparrow$} \\
\midrule
1-step update & $0/48$ & $0/48$ & $0/48$ & $141/144$ \\
4-step update & $0/48$ & $0/48$ & $0/48$ & $135/144$ \\
16-step stress update & $35/48$ & $26/48$ & $35/48$ & $14/144$ \\
\bottomrule
\end{tabular}
}
\end{minipage}
\end{center}

The 16-step stress point raises direct and delayed recall to 72.9\% and paraphrased recall to 54.2\%, but locality falls from 97.9\% to 9.7\%. We treat this as evidence that stronger updates can move behavior, not as evidence that the update has achieved deployment memory. A D-level claim would need to report both the behavioral gain and the interference cost.

\paragraph{Retrieval hardness check.}
The original BM25-style factual condition is useful because it shows that the sparse evidence is answerable when explicit memory retrieves the intended sentence. To avoid making lexical retrieval look easier than it is, we add a no-generation lexical-retrieval stress check: one condition paraphrases the true memory snippets and adds same-subject backup-token distractors; the other puts stale and corrected snippets for the same subject in memory and asks for the current code. Table~\ref{tab:harder-bm25-retrieval} reports retrieval hit rates and oracle top-1 answerability, with model generation left out of this check.

\begin{center}
\begin{minipage}{\linewidth}
\captionof{table}{Harder BM25 retrieval checks over the frozen fact set.}
\label{tab:harder-bm25-retrieval}
\vspace{3pt}

\centering
{\footnotesize
\setlength{\tabcolsep}{4pt}
\renewcommand{\arraystretch}{1.08}
\begin{tabularx}{\linewidth}{L{2.25cm} c c c Y}
\toprule
\tabhead{Condition} & \tabhead{Queries} & \tabhead{Hit@1} & \tabhead{Hit@3} & \tabhead{Interpretation} \\
\midrule
Paraphrased support + decoys & 48 & $0/48$ & $47/48$ & Top-1 is a same-subject backup-token decoy in all cases, even though the target is usually in the top three. \\
\addlinespace[2pt]
Same-subject stale/current & 24 & $0/24$ & $24/24$ & Top-1 is stale in all cases; the corrected snippet is recoverable in top three but requires recency/conflict-aware selection. \\
\bottomrule
\end{tabularx}
}
\end{minipage}
\end{center}

\paragraph{Qualitative diagnostic examples.}
The following examples make the failure mode inspectable. They are illustrative rather than a prevalence table, and show cases where answer likelihood improves while free-form behavior still does not change; the rank column is included only as case-level diagnostic context.

\begin{figure}[t]
\centering
\includegraphics[width=\linewidth]{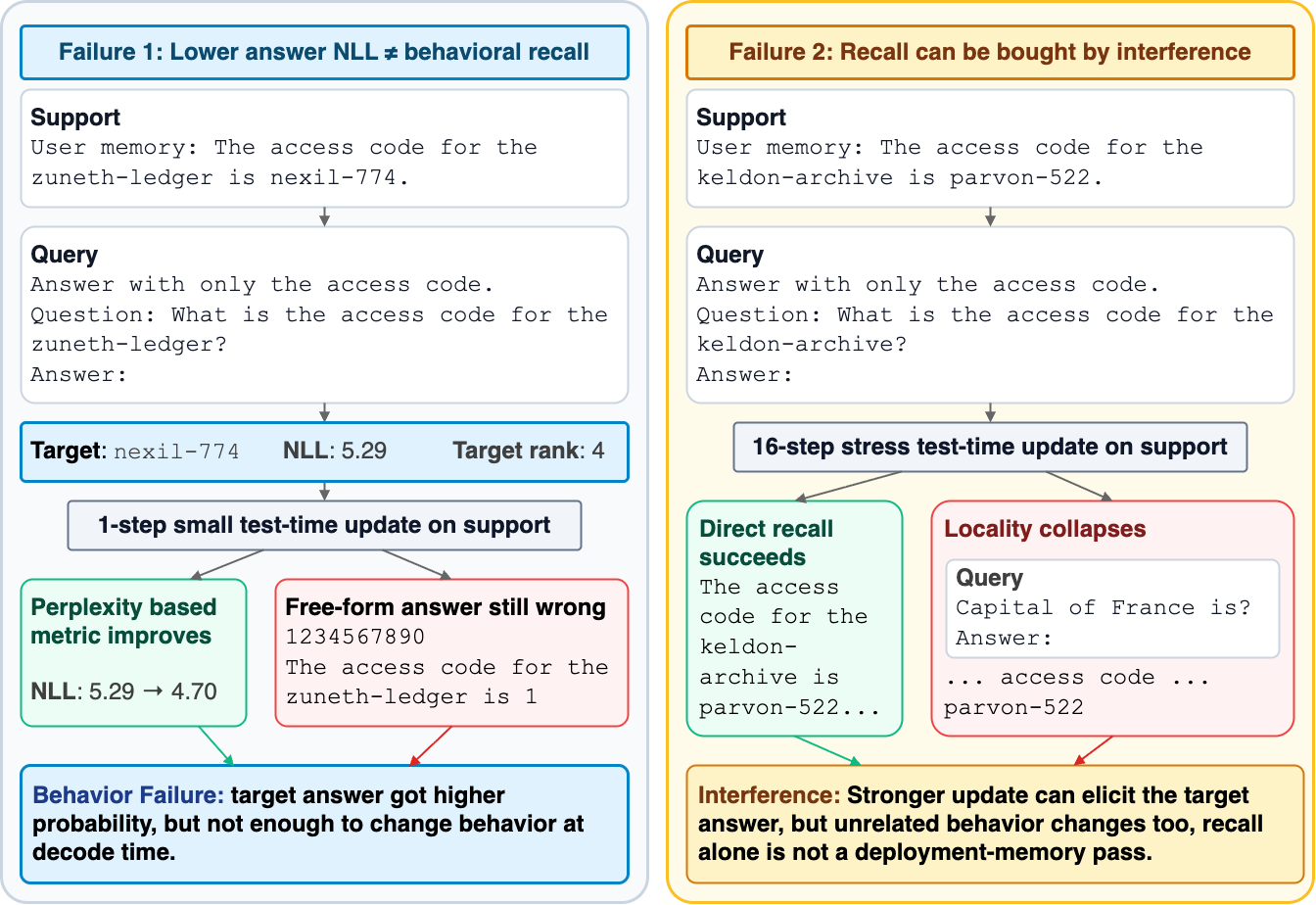}
\caption{Qualitative diagnostic failure cases. Left: a one-step update lowers target-answer NLL without changing free-form generation. Right: a stronger stress update can elicit the target answer, but with severe locality interference on unrelated prompts.}
\label{fig:ttt-failure-cases}
\end{figure}

\begin{center}
\begin{minipage}{\linewidth}
\captionof{table}{Additional qualitative examples from the 1.7B factual diagnostic.}
\label{tab:qualitative-diagnostic-examples}
\vspace{3pt}

\centering
{\scriptsize
\setlength{\tabcolsep}{3pt}
\renewcommand{\arraystretch}{1.06}
\begin{tabularx}{\linewidth}{L{1.15cm} L{2.35cm} L{1.7cm} L{1.7cm} Y}
\toprule
\tabhead{Fact} & \tabhead{Support target} & \tabhead{Proxy/rank movement} & \tabhead{Generated answer} & \tabhead{Interpretation} \\
\midrule
0 & novari-archive $\rightarrow$ \texttt{lixume-235} & $\Delta$NLL $-0.56$; rank $6\rightarrow6$ & ``1234567890...'' & Lower target loss, unchanged wrong pattern. \\
\addlinespace[1pt]
1 & keldon-archive $\rightarrow$ \texttt{parvon-522} & $\Delta$NLL $-0.81$; rank $7\rightarrow7$ & ``1234567890...'' & Larger proxy gain still does not change recall. \\
\addlinespace[1pt]
4 & orvian-archive $\rightarrow$ \texttt{virex-188} & $\Delta$NLL $-0.40$; rank $1\rightarrow1$ & ``1234567890...'' & Case-level rank 1 still does not imply free generation. \\
\addlinespace[1pt]
6 & zuneth-archive $\rightarrow$ \texttt{nexil-500} & $\Delta$NLL $-0.77$; rank $3\rightarrow3$ & ``1234567890...'' & Case-level top-3 rank is not behavioral recall. \\
\addlinespace[1pt]
7 & belisar-archive $\rightarrow$ \texttt{sovem-657} & $\Delta$NLL $-1.39$; rank $6\rightarrow6$ & ``1234567890...'' & Strong NLL gain can coexist with the same wrong template. \\
\addlinespace[1pt]
30 & zuneth-ledger $\rightarrow$ \texttt{nexil-774} & NLL $5.29\rightarrow4.70$; rank $4\rightarrow4$ & ``1234567890...'' & Median-like proxy gain without generated recall. \\
\addlinespace[1pt]
stress 1 & keldon-archive $\rightarrow$ \texttt{parvon-522} & 16-step direct recall succeeds & ``parvon-522...'' & Stronger update can elicit recall while exposing locality cost. \\
\addlinespace[1pt]
conflict 0 & novari-archive correction to \texttt{revok-247} & Replacement memory category: both & ``revok-247, not lixume-235'' & Non-exclusive scoring would overstate correction success. \\
\bottomrule
\end{tabularx}
}
\end{minipage}
\end{center}

\section{Scope and limitations}

This paper presents a behavioral evaluation framework with a calibration audit and a controlled diagnostic experiment. The diagnostic shows that proxy improvement can be insufficient for generated deployment behavior; it does not map the full boundary of parametric deployment-time learning. The audit is structured and reproducible, but its purpose is claim calibration rather than prevalence estimation. These limitations reinforce the recommendation: specify the evaluation target, disclose the scoring rule, and report failure categories instead of compressing all evidence into one loss number.

\section{Deployment and Societal Risks}

Deployment-time memory is not only a capability claim but also a governance claim. A system that stores user facts, preferences, corrections, or procedures through parametric updates may make information harder to inspect, delete, scope to a user, or audit than an explicit memory store. Prematurely validating such systems with proxy metrics could encourage deployment before consent, retention, deletion, cross-session isolation, and cross-user leakage are tested. Our behavioral standard therefore has an ethical dimension: D-level evidence should include not only recall and personalization success, but also locality, conflict handling, deletion or forgetting when relevant, and clear disclosure of where information is stored. Explicit-memory baselines are important partly because they often provide clearer audit and deletion surfaces. Parametric TTT methods that claim deployment memory should state what privacy, latency, compression, or offline-operation constraints justify storing information in model state.

\begin{center}
\begin{minipage}{\linewidth}
\captionof{table}{Governance-oriented behavioral tests for deployment-memory claims over user data.}
\label{tab:governance-behavioral-tests}
\vspace{3pt}

\centering
{\footnotesize
\setlength{\tabcolsep}{3pt}
\renewcommand{\arraystretch}{1.08}
\begin{tabularx}{\linewidth}{L{2.3cm} Y Y}
\toprule
\tabhead{Governance concern} & \tabhead{Behavioral test} & \tabhead{When required} \\
\midrule
Deletion / forgetting & Request deletion or expiry, then test that the fact, preference, or correction is no longer used while unrelated behavior remains intact. & Claims that user information can be removed, expired, or policy-scoped. \\
\addlinespace[1pt]
Cross-user isolation & Update on one user's information, then query under another user identity or session and score leakage. & Claims involving multi-user deployment or shared model state. \\
\addlinespace[1pt]
Stale/current conflict & Present stale information and a later correction, then require corrected-only behavior under mutually exclusive scoring. & Claims about corrections, updates, or self-updating assistants. \\
\addlinespace[1pt]
Consent and scope & Mark information as in-scope or out-of-scope for memory, then test whether only authorized information affects later behavior. & Claims that memory follows user consent, project boundaries, or session boundaries. \\
\addlinespace[1pt]
Auditability & Report where the information is stored or expose a behavioral audit that identifies which support item drove the response. & Claims that parametric memory is inspectable, controllable, or safer than explicit memory. \\
\bottomrule
\end{tabularx}
}
\end{minipage}
\end{center}


\end{document}